# Solving POMDPs by Searching in Policy Space

**Eric A. Hansen**
Computer Science Department
University of Massachusetts
Amherst, MA 01003
hansen@cs.umass.edu

## Abstract

Most algorithms for solving POMDPs iteratively improve a value function that implicitly represents a policy and are said to search in value function space. This paper presents an approach to solving POMDPs that represents a policy explicitly as a finite-state controller and iteratively improves the controller by search in policy space. Two related algorithms illustrate this approach. The first is a policy iteration algorithm that can outperform value iteration in solving infinite-horizon POMDPs. It provides the foundation for a new heuristic search algorithm that promises further speedup by focusing computational effort on regions of the problem space that are reachable, or likely to be reached, from a start state.

## 1   Introduction

A partially observable Markov decision process (POMDP) provides an elegant mathematical model for planning and control problems for which there can be uncertainty about the effects of actions and about the current state. It is well-known that a state probability distribution updated by Bayesian reasoning is a sufficient statistic that summarizes all information about the history of the process necessary for optimal action selection. Therefore the standard approach to solving a POMDP is to recast it as a completely observable MDP with a state space that consists of all possible state probability distributions. In this form, it is solved using dynamic programming or related techniques that rely on the Markov assumption.

Algorithms for solving POMDPs in this way rely on a value function that maps state probability distributions to expected values. A value function defined for all possible state probability distributions can be represented in different ways; for example, as a set of vectors and a max operator (Smallwood & Sondik 1973) or as a grid of point values with an interpolation rule (*e.g.*, Hauskrecht 1997). Given some explicit representation of the value function, a policy is represented implicitly by the same value function and one-step lookahead. Most algorithms for solving POMDPs represent a policy implicitly in this way and improve the policy by gradually improving the value function, typically by repeated "backups" using value iteration or reinforcement learning. Because the policy is only represented implicitly by the value function, such algorithms are said to search in value function space.

This paper presents an approach to solving POMDPs that represents a policy explicitly and relies on search in policy space. In this approach, choice of how to represent a policy is critical in a way that it is not for algorithms that search in value function space. It is possible to represent a policy explicitly as a mapping from state probability distributions to actions by partitioning probability space into a finite set of regions and mapping each region to some action. Sondik (1978) describes a policy iteration algorithm that represents a policy in this way. However this algorithm is very complex and difficult to implement and, as a result, is not used in practice.

In this paper, we consider an alternative representation of a policy as a finite-state controller and present two related algorithms for solving infinite-horizon POMDPs by searching in a policy space of finite-state controllers. The first is a policy iteration algorithm, first described by Hansen (1998a), that simplifies policy iteration for POMDPs by representing a policy as a finite-state controller. It provides the foundation for a related heuristic search algorithm, presented here for the first time, that can focus computational effort on regions of the search space that are reachable, or likely to be reached, from a given start state.

## 2  Background

Consider a discrete-time POMDP with a finite set of states $S$, a finite set of actions $A$, and a finite set of observations $Z$. Each time period, the system is in some state $s \in S$, an agent chooses an action $a \in A$ for which it receives an immediate reward with expected value $r(s, a) \in \Re$, the system makes a transition to state $s' \in S$ with probability $Pr(s'|s, a) \in [0, 1]$, and the agent observes $z \in Z$ with probability $Pr(z|s', a) \in [0, 1]$. The state of the system cannot be directly observed, but the probability that it is in a given state can be calculated. Let $b$ denote a vector of state probabilities, called a *belief state*, where $b(s)$ denotes the probability that the system is in state $s$. If action $a$ is taken and observation $z$ follows, the successor belief state, denoted $b_z^a$, is determined by revising each state probability as follows,

$$b_z^a(s') = \frac{Pr(z|s', a) \sum_{s \in S} Pr(s'|s, a) b(s)}{Pr(z|b, a)},$$

where the denominator is a normalizing factor $Pr(z|b, a) = \sum_{s' \in S} Pr(z|s', a) \sum_{s \in S} Pr(s'|s, a) b(s)$.

A POMDP is solved by finding a rule for selecting actions, called a policy, that optimizes a performance objective (or comes acceptably close to doing so). We assume the objective is to maximize the expected total discounted reward over an infinite horizon (where $\beta \in (0, 1]$ is a discount factor). By recasting a POMDP as a completely observable MDP with a continuous, $|S|$-dimensional state space that consists of all possible belief states, the problem can be solved by iteration of a *dynamic-programming update* that performs the following "one-step backup" for each belief state $b$:

$$V'(b) := \max_{a \in A} \left[ \sum_{s \in S} b(s) r(s, a) + \beta \sum_{z \in Z} Pr(z|b, a) V(b_z^a) \right]. \tag{1}$$

In words, this says that the value of belief state $b$ is set equal to the immediate reward for taking the best action for $b$ plus the discounted expected value of the resulting belief state $b_z^a$. Iteration of the dynamic-programming update, called *value iteration*, converges to the optimal value function in the limit. However the number of belief states that must be "backed-up" each iteration is uncountably infinite and it is not obvious how to do this.

The key to computing the dynamic-programming update is Smallwood and Sondik's (1973) proof that it preserves the piecewise linearity and convexity of the value function. A piecewise linear and convex value function $V$ can be represented by a finite set of $|S|$-dimensional vectors of real numbers, $\mathcal{V} = \{v^0, v^1, \ldots, v^k\}$, such that the value of each belief state is defined as follows:

$$V(b) = \max_{0 \leq i \leq k} \sum_{s \in S} b(s) v^i(s).$$

The dynamic-programming update transforms a value function $V$ represented in this way into an improved value function $V'$ represented by another finite set of vectors, $\mathcal{V}'$. Several algorithms for performing the dynamic-programming update have been developed. All rely heavily on linear programming and are computationally intensive; the algorithm that is presently the fastest is described by Cassandra, Littman and Zhang (1997). We do not describe here how to compute the dynamic-programming update and instead refer to this paper, Kaelbling *et al.* (1996), Cassandra *et al.* (1994), and references therein.

Algorithms that search in value function space, such as value iteration, must be able to extract a policy from the value function they iteratively improve. There are two possible ways to do so that correspond to two possible representations of a policy.

One possibility is to view a policy as a mapping from belief states to actions. Given some representation of a value function mapping belief states to values, a policy $\delta$ is extracted using one-step lookahead,

$$\delta(b) = \arg \max_{a \in A} \left[ \rho(b, a) + \beta \sum_{z \in Z} Pr(z|b, a) V(b_z^a) \right], \tag{2}$$

where $\rho(b, a) = \sum_{s \in S} b(s) r(s, a)$ is the expected immediate reward for taking action $a$ in belief state $b$.

A second possibility is to represent a policy as a finite-state controller. A correspondence between vectors and one-step policy choices plays an important role in this interpretation of a policy. Each vector in $\mathcal{V}'$ corresponds to the choice of an action, and for each possible observation, choice of a vector in $\mathcal{V}$. Among all possible one-step policy choices, the vectors in $\mathcal{V}'$ correspond to those that optimize the value of some belief state. To describe this correspondence between vectors and one-step policy choices, we introduce the following notation. For each vector $v^i$ in $\mathcal{V}'$, let $a(i)$ denote the choice of action and, for each possible observation $z$, let $l(i, z)$ denote the index of the successor vector in $\mathcal{V}$. Given this correspondence between vectors and one-step policy choices, Kaelbling *et al.* (1996) point out that an optimal policy for a finite-horizon POMDP can be represented by an acyclic finite-state controller in which each machine state corresponds to a vector in a nonstationary value function.

Value iteration can also be used to solve infinite-horizon POMDPs. The optimal value function for an infinite-horizon POMDP is not necessarily piecewise linear, although it is convex. However it can be approximated arbitrarily closely by a piecewise linear and convex function. Moreover Sondik (1978) and Cassandra *et al.* (1994) point out that sometimes, although not reliably, value iteration converges to an optimal piecewise linear and convex value function that is equivalent to a cyclic finite-state controller. The finite-state controller can be extracted from the value function using the correspondence between vectors and one-step policy choices noted earlier. However a finite-state controller cannot reliably be extracted from a suboptimal value function and so a policy for an infinite-horizon POMDP is generally viewed as a mapping from belief states to actions represented implicitly by a value function and extracted using equation (2).

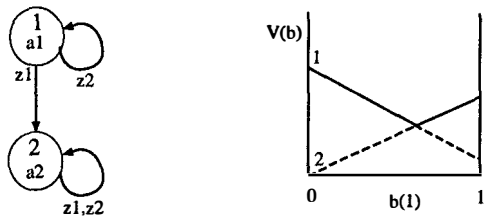

Figure 1: Example of a simple finite-state controller and corresponding value function for a POMDP with two states, two actions (a1,a2) and two observations (z1,z2). Each machine state is labeled by a unique number and by an action to take in that state; state transitions are labeled by observations. Each vector of the value function is labeled by the number of the machine state to which it corresponds. For a two-state POMDP, the belief state can be represented by a single real number between 0 and 1 that represents the probability of being in one of the states; the horizontal axis in the figure on the right represents the belief state in this way. The vertical axes represent the expected value of a belief state. The value function is the upper surface of the vectors and reflects the rule that the controller is started in the machine state that optimizes the value of the starting belief state.

For algorithms that search in value function space, it is important to be able to extract a policy from a value function. For algorithms that search in policy space, it is equally important to be able to compute the value function of a policy; this is called policy evaluation. We conclude this review by pointing out that for a policy represented as a finite-state controller, policy evaluation is straightforward. A piecewise linear and convex value function can be computed by solving the following system of linear equations, where there is one equation for each pair of machine state $i$ and system state $s$:

$$v^i(s) = r(s, a(i)) + \qquad\qquad (3)$$
$$\beta \sum_{s', z} Pr(s'|s, a(i)) Pr(z|s', a(i)) v^{l(i, z)}(s').$$

The value function has one linear facet or $|S|$-vector for each machine state of the finite-state controller. Al-though the policy is a finite-state controller, the value function is defined for belief space and the controller is started in the machine state that corresponds to the vector that optimizes the value of the starting belief state. (See Figure 1.)

## 3   Policy Iteration

The first algorithm we consider that solves a POMDP by searching in policy space is policy iteration. Because it includes a policy evaluation step that computes the value function of a given policy, it must represent the policy explicitly and independently of the value function. Sondik (1978) describes a policy iteration algorithm for POMDPs that represents a policy as a mapping from a finite number of polyhedral regions of belief space to actions. Each region of belief space is represented by a set of linear inequalities that define its boundaries. Because there is no known method for computing the value function of a policy represented in this way, the policy evaluation step of Sondik's algorithm converts a policy from this representation to an equivalent, or approximately equivalent, finite-state controller; as we have seen, the value function of a finite-state controller can be computed in a straightforward way. However conversion between these two representations is extremely complicated and difficult to implement. As a result, Sondik's algorithm is not used in practice.

We now show that policy iteration for POMDPs can be simplified by representing a policy as a finite-state controller. The obvious simplification is that this makes policy evaluation, the most difficult step of Sondik's algorithm, straightforward. But for this approach to work, we must show that the dynamic-programming update can be interpreted as the transformation of a finite-state controller $\delta$ into an improved finite-state controller $\delta'$; that is, we must show how to perform policy improvement on finite-state controllers. We do this by showing that a simple comparison of the vectors in $\mathcal{V}^\delta$ and $\mathcal{V}'$ provides the basis for such a transformation, where $\mathcal{V}^\delta$ is the set of vectors that represents the value function of the current finite-state controller $\delta$ and $\mathcal{V}'$ is the output of the dynamic-programming update given $\mathcal{V}^\delta$ as input.

First recall that every vector $v^i$ in $\mathcal{V}^\delta$ is associated with an action, denoted $a(i)$, and for each possible observation $z$, a transition to another vector in $\mathcal{V}^\delta$, with index $l(i, z)$. This follows from the fact that $\mathcal{V}^\delta$ is computed by evaluating a finite-state controller. Similarly every vector $v^j$ in $\mathcal{V}'$ found by the dynamic-programming update is associated with an action, $a(j)$, and for each possible observation $z$, a transition to a vector in $\mathcal{V}^\delta$, where $l(j, z)$ denotes the index of the vector.

1. Specify an initial finite-state controller, $\delta$, and select $\epsilon$ for detecting convergence to an $\epsilon$-optimal policy.

2. Policy evaluation: Compute the value function for $\delta$ by solving the system of equations given by equation (4).

3. Policy improvement:

   (a) Perform a dynamic-programming update that transforms a set of vectors $\mathcal{V}^\delta$ into a set of vectors $\mathcal{V}'$.

   (b) For each vector $v^i$ in $\mathcal{V}'$:

      i. If the action and successor links associated with it are the same as those of a machine state already in $\delta$, then keep that machine state unchanged in $\delta'$.

      ii. Else if the vector $v^i$ pointwise dominates a vector associated with a machine state of $\delta$, *change* the action and successor links of that machine state to those that correspond to $v^i$. (If it pointwise dominates the vectors of more than one machine state, they can be combined into a single machine state.)

      iii. Else *add* a machine state to $\delta'$ that has the action and successor links associated with $v^i$.

   (c) *Prune* any machine state of $\delta'$ for which there is no corresponding vector in $\mathcal{V}'$, as long as it is not reachable from a machine state to which a vector in $\mathcal{V}'$ does correspond.

4. Termination test. Calculate the Bellman residual and if it is less than or equal to $\epsilon(1-\beta)/\beta$, exit with an $\epsilon$-optimal policy. Otherwise set $\delta$ to $\delta'$. If some node was changed in step (3b), goto step 2; otherwise goto step 3.

Figure 2: Policy iteration algorithm for POMDPs.

Vectors in $\mathcal{V}'$ can be duplicates of vectors in $\mathcal{V}^\delta$, that is, they can have the same action and successor links (in which case their vector values will be pointwise equal). If they are not duplicates, they indicate how the finite-state controller can be changed to improve the value function – either by changing a machine state (that is, changing its corresponding action and/or successor links) or by adding a machine state. There may also be some machine states for which there is no corresponding vector in $\mathcal{V}'$ and they can be *pruned*, but only if they are not reachable from a machine state that corresponds to a vector in $\mathcal{V}'$. (This last point is important because it preserves the integrity of the finite-state controller.) Thus a finite-state controller can be iteratively improved using a combination of three transformations; changing machine states, adding machine states, and pruning machine states.

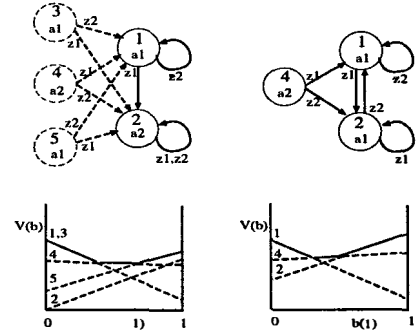

Figure 3: Example of how the finite-state controller of Figure 1 can be improved by a step of policy iteration. The dynamic-programming update returns vectors 3, 4 and 5 which have corresponding potential machine states shown in the left panel as dashed circles. Node 3 is a duplicate of machine state 1 and causes no change. Node 4 becomes a new machine state. The vector for node 5 pointwise dominates the vector for machine state 2 and therefore the action and observation links for machine state 2 can be changed accordingly. The improved finite-state controller and its value function are shown in the right panel.

Figure 2 outlines a policy iteration algorithm with a policy improvement step that uses these simple transformations to improve a finite-state controller. Figure 3 illustrates the policy improvement step with a simple example. Transformation of the finite-state controller after performing the dynamic-programming update adds little overhead to the policy improvement step because it simply compares the vectors in $\mathcal{V}'$ to the vectors in $\mathcal{V}^\delta$ and modifies the finite-state controller accordingly. If a machine state is changed, the policy evaluation step is invoked to compute the value function of the transformed finite-state controller. We can prove the following generalization of Howards's policy improvement theorem (Hansen 1998b).

**Theorem 1** *If a finite-state controller is not optimal, policy improvement transforms it into a finite-state controller with a value function that is as good or better for every belief state and better for some belief state.*

If the policy improvement step does not change the finite-state controller, that is, if all the vectors in $\mathcal{V}'$ are duplicates of vectors in $\mathcal{V}^\delta$, then the Bellman optimality equation is satisfied and the finite-state controller must be optimal. Therefore policy iteration can detect convergence to an optimal finite-state controller. However not every POMDP has an optimal finite-state controller and policy iteration may simply find a succession of finite-state controllers that are increasingly close approximations of an optimal policy. We use the same stopping condition Sondik uses to detect $\epsilon$-optimality: a finite-state controller is $\epsilon$-optimal when the Bellman residual is less than or equal to $\epsilon(1 - \beta)/\beta$, where $\beta$ is the discount factor, and we can prove the following convergence result (Hansen 1998b).

Table 1: Comparison of value iteration and policy iteration on nine test problems from Cassandra *et al.* (1997). The last four columns show the number of CPU seconds until convergence to $\epsilon$-optimality for each of four values of $\epsilon$; 10.0, 1.0, 0.1, and 0.01. For each test problem, the timing results for value iteration are shown above the timing results for policy iteration. The algorithm was sometimes terminated before reaching $\epsilon = 0.01$.

| Test problem | 10.0 | 1.0 | 0.1 | 0.01 |
|---|---|---|---|---|
|              | < 1   | < 1   | 2     | 3     |
| 1D maze      | < 1   | < 1   | < 1   | < 1   |
|              | 3     | 29    | 100   | 171   |
| 4x3CO        | 1     | 3     | 3     | 3     |
|              | 2     | 9251  | 61973 |       |
| 4x3          | 1     | 868   | 4951  | 10935 |
|              | 13    | 93    | 275   | 457   |
| 4x4          | 3     | 17    | 17    | 17    |
|              | 10    | 103   | 305   | 512   |
| Cheese       | 12    | 12    | 12    | 12    |
|              | 2     | 1363  | 1776  | 1852  |
| Part painting| 2     | 10    | 31    | 31    |
|              | 10735 | 19557 | 28289 | 37061 |
| Network      | 259   | 1656  | 2239  | 3132  |
|              | 4346  | 7545  | 10882 | 14258 |
| Shuttle      | 78    | 151   | 245   | 340   |
|              | 61472 | 273678 |      |       |
| Aircraft ID  | 772   | 11548 | 91234 |       |

**Theorem 2** *Policy iteration converges to an $\epsilon$-optimal finite-state controller after a finite number of iterations.*

As with completely observable MDPs, policy iteration can converge to $\epsilon$-optimality (or optimality) in fewer iterations than value iteration because interleaving a policy evaluation step with the dynamic-programming update accelerates improvement of the value function. For completely observable MDPs, this is not a clear advantage because the policy evaluation step is more computationally expensive than the dynamic-programming update. For POMDPs, policy evaluation has low-order polynomial complexity compared to the worst-case exponential complexity of the dynamic-programming update (Littman *et al.* 1995). Therefore, policy iteration appears to have a clearer advantage over value iteration for POMDPs.

Table 1 compares the performance of value iteration and policy iteration on nine test problems from Cassandra *et al.* (1997). (For these problems, the average number of states is 9.3, the average number of actions is 3.9, and the average number of observations is 4.7.) Their incremental pruning algorithm was used to perform dynamic-programming updates in both value iteration and policy iteration and experiments were performed on a AlphaStation 200/4 with a 233Mhz processor and 128M of RAM. The results show that policy iteration consistently outperforms value iteration and the increased rate of convergence is often dramatic.

A few of these test problems have small optimal finite-state controllers (1D maze, 4x3CO, 4x4, Cheese and part painting). For them, policy iteration converges quickly and sometimes reduces the error bound from more than 1.0 to zero in a single iteration. For the other problems, finite-state controllers with between a couple hundred and several hundred machine states are generated without converging to optimality. This illustrates that the difficulty of solving a POMDP is primarily a function of the size of the controller needed to achieve good performance and not simply a function of the number of states, actions and observations.

## 4   Heuristic search

Although policy iteration converges more quickly that value iteration, both are limited to solving very small POMDPs. The shared bottleneck is the dynamic-programming update. The fastest algorithm for performing it is still prohibitively slow for problems with more than about ten or fifteen states, actions, or observations. Policy iteration is faster that value iteration because it takes fewer iterations of the dynamic-programming update to converge. But when a single iteration is computationally prohibitive, policy iteration is as impractical as value iteration. In this section, we introduce a new approach to solving POMDPs that is closely related to the policy iteration algorithm described in the previous section but differs in an important respect; it does not use the dynamic-programming update to improve a policy. Instead it uses heuristic search.

Heuristic search has been used before to solve POMDPs approximately. Satia and Lave (1973) describe a branch-and-bound algorithm for solving infinite-horizon POMDPs, given an initial belief state, and Larsen (1989) and Washington (1996,1997) use the best-first heuristic search algorithm AO* in a similar way. For infinite-horizon problems, it is only possible to search to a finite depth and these algorithms find a solution that takes the form of a tree that grows with the depth of the search. The search tree can be repre-sented by an AND/OR tree in which the nodes of the tree correspond to belief states and the root of the tree is the initial belief state. An OR node represents the choice of an action and an AND node represent a set of possible observations. The value of an OR node is the value of the best action for the belief state that corresponds to it. The value of an AND node is the sum of the values of the belief states that follow each observation, multiplied by the probability of each observation. Upper and lower bounds are computed for belief states on the fringe of the search tree and backed-up through the tree to the starting belief state at its root. Thus expanding the search tree improves the bounds at the interior nodes of the tree. The error bound (the difference between the upper and lower bounds on the value of the starting belief state) can be made arbitrarily small by expanding the search tree far enough and, for discounted POMDPs, an ε-optimal policy for the belief state at the root of the tree can be found after a finite search (Satia and Lave 1973).

Several possible upper bound functions for evaluating the fringe nodes of the search tree have been discussed by others (e.g., Hauskrecht 1997, Brafman 1997) and we do not add to that discussion here. For a lower bound function, we use the piecewise linear and convex value function of a finite-state controller and improve the lower bound during search by iteratively improving the finite-state controller, much as policy iteration does. This is the principal innovation of our heuristic search algorithm.

Recall that every node of the search tree corresponds to a belief state. Therefore expanding an OR node (and all its child AND nodes), and backing up its lower bound, is equivalent to performing a one-step backup for the corresponding belief state (as in equation 1). This backup may improve the lower bound of the belief state and, if it does, we know that a machine state can be added to the finite-state controller that improves the value of at least this one belief state. Therefore expanding a search node performs a similar function as the dynamic-programming update, and can be interpreted in a similar way as the (potential) modification of a finite-state controller; the difference is that a node expansion corresponds to a one-step backup for a single belief state whereas the dynamic-programming update performs a one-step backup for all possible belief states.

When the lower bound for a belief state in the search tree is improved, it is backed-up through the search tree and possibly improves the lower bound of the starting belief state at the root. When it does so, the search algorithm has found a way to improve the value of the starting belief state by modifying the finite-state controller. The finite-state controller is modified as

1. Specify an initial finite-state controller, $\delta$, and select $\epsilon$ for detecting convergence to an $\epsilon$-optimal policy.

2. Policy evaluation: Compute the value funtion for $\delta$ by solving the system of equations given by equation (4).

3. Policy improvement;

   (a) Perform forward search from the starting belief state and back up lower and upper bounds from the leaves of the search tree. Continue until either the lower bound of the starting belief state is improved or the error bound on the value of the starting belief state is less than or equal to $\epsilon$.

   (b) If the error bound is less than or equal to $\epsilon$, exit with an $\epsilon$-optimal policy. Otherwise continue.

   (c) Given that forward search has found a change of policy that improves the lower bound of the starting belief state, consider every reachable node in the search tree for which the lower bound has been improved. (A node is said to be reachable if it can be reached by starting from the root node and always selecting actions that optimize the lower bound). For each of these nodes in order from the leaves to the root:

      i. If its action and successor links are the same as those of a machine state of $\delta$, then keep that machine state unchanged in $\delta'$.

      ii. Else compute the vector for this node and if it pointwise dominates the vector for a machine state of $\delta$, *change* the action and successor links of that machine state to those of this node. (If it pointwise dominates the vectors of more than one machine state, they can be combined into a single machine state.)

      iii. Else *add* a machine state to $\delta'$ that has the same action and successor links as this node.

   (d) *Prune* any machine state of $\delta'$ that is not reachable from the machine state that optimizes the value of the starting belief state.

4. Set $\delta$ to $\delta'$. If some machine state of the controller has been changed in (3c), goto step 2; otherwise goto step 3.

Figure 4: Heuristic search algorithm for POMDPs.

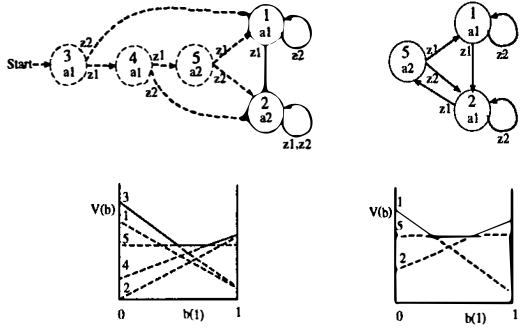

Figure 5: Example of how the finite-state controller of Figure 1 can be improved by heuristic search. Nodes 3, 4 and 5 in the left panel are potential new machine states that correspond to a path through the search tree, beginning from the starting belief state, for which the lower bound value of each belief state has been improved. Node 5 becomes a new machine state. The vector corresponding to node 4 pointwise dominates the vector for machine state 2 and therefore the action and observation links for machine state 2 can be changed accordingly. Node 3 is a duplicate of machine state 1 and causes no change. The improved finite-state controller and its value function are shown in the right panel.

follows. Beginning at the root of the search tree and selecting the action that optimizes the lower bound of each OR node, each reachable node of the search tree for which the lower bound has been improved is identified. For each of these nodes, in backwards order from the fringe of the search tree, a corresponding vector is computed based on the current value function and the finite-state controller is modified using the same transformations used by the policy iteration algorithm. The algorithm is summarized in Figure 4 and illustrated by a simple example in Figure 5.

Heuristic search can recognize when to change machine states (by detecting pointwise dominance) as well as when to add them. When a machine state is changed, policy evaluation is invoked to recompute the value function; like policy iteration, this algorithm interleaves a policy improvement step with a policy evaluation step. Any machine state that is not reachable from the machine state that optimizes the value of the starting belief state can be pruned without affecting the value of the starting belief state and the error bound is simply the difference between the upper and lower bounds for the starting belief state.

In our implementation, we use AO* to perform heuristic search. It expands the search tree in a best-first order and we use the upper bound function to identify the most promising solution tree. To select one of its fringe nodes for expansion, we use the following heuristic: select for expansion the node (and corresponding

belief state $b$) for which the value

$$(UB(b) - V(b)) * (REACHPROB(b) * \beta^{DEPTH(b)})$$

is greatest, where $UB$ denotes the upper bound function, the lower bound function is the the value function $V$ of the current finite-state controller, $REACHPROB(b)$ denotes the probability of reaching belief state $b$ beginning from the starting belief state at the root of the tree, and $DEPTH(b)$ denotes the depth of belief state $b$ in the search tree (measured by the number of actions taken along a path from the root). This selection heuristic focuses computational effort where it is most likely to improve the bounds of the starting belief state.

There are several advantages to using heuristic search instead of the dynamic-programming update to improve a finite-state controller. First and most importantly, it adds machine states to the finite-state controller only if they improve the value of the starting belief state. By contrast, policy iteration finds a finite-state controller that optimizes the value of every possible belief state and this is usually a much larger controller. Among the test problems listed in Table 1, for example, an optimal finite-state controller for the cheese grid problem has five machine states when the starting belief state is a uniform probability distribution; policy iteration converges to a finite-state controller with fourteen machine states that optimizes all possible starting belief states. A related advantage of heuristic search is that it can focus computational effort on regions of belief space that are likely to be reached from the starting belief state; for example, it can focus search on the most probable trajectores through the search tree. It also avoids use of linear programming, the most computationally intensive part of the dynamic-programming update.

The theoretical properties of the algorithm are similar to those for policy iteration, but are specialized to a starting belief state (Hansen 1998b).

**Theorem 3** *If a finite-state controller does not optimize the value of the starting belief state, heuristic search transforms it into a finite-state controller with an improved value for the starting belief state.*

**Theorem 4** *The heuristic search algorithm converges after a finite number of steps to a finite-state controller that is ε-optimal for the starting belief state.*

For the test problems of Table 1 and several other small POMDPs, this heuristic search algorithm improves the value of the finite-state controller for a starting belief state faster than policy iteration, and often considerably faster. Results are mixed for improvement of the error bound. For some problems the error bound converges quickly to zero or close to it; for others it converges more slowly and the AO* search algorithm used in the policy improvement step runs out of memory trying to reduce it further. How quickly the error bound converges depends primarily on how closely the upper bound function estimates the optimal value function for a particular problem. If the upper bound function is not a good estimate, the error bound can only be improved by deep expansion of the search tree. Because the quality of the upper bound also determines how aggressively the search tree can be pruned, a poor upper bound function can cause the size of the search tree to quickly exceed available memory. Sophisticated methods for computing upper bound functions have been developed that we have not yet implemented (e.g., Hauskrecht 1997; Brafman 1997) and we expect these will improve performance of the heuristic search algorithm and accelerate convergence of the error bound. We also plan to implement a memory-bounded version of AO* that can search more deeply in the tree (Chakrabarti et al. 1990; Washington 1997).

The most promising aspect of this heuristic search algorithm is its potential for solving problems for which the dynamic-programming update is computationally prohibitive. Consider a simple maze problem described by Hauskrecht (1997) that has 20 states, 6 actions, and 8 observations. Although still a very small problem, it is out of the range of dynamic programming. We tested policy iteration on this problem with an initial finite-state controller with a single machine state. The first iteration of policy iteration took a fraction of a second and resulted in a improved finite-state controller with five machine states. The second iteration took two minutes and resulted in a improved finite-state controller with 172 machine states. In the third iteration, the dynamic-programming update ran for 20 hours without finishing, at which point policy iteration was terminated. Clearly this is a problem for which dynamic programming seems computationally prohibitive. The finite-state controller found after two iterations (and two minutes) had an error bound of 539.7 and a value of 34.3 for a starting belief state that is a uniform state probability distribution. On the same maze problem, our heuristic search algorithm found a finite-state controller with 96 machine states and a value of 52.2 for the same starting belief state, a significant improvement in performance achieved by a smaller controller. After several minutes of expanding the search tree, the algorithm ran out of memory after reducing the error bound to 25.4.

Of course, this is only a single example and our results are preliminary. But it does at least suggest this heuristic search algorithm may extend the range of problems to which the policy space approach described in this paper can be applied. It may do so for several reasons; it eliminates the need to perform the dynamic-programming update, it improves a finite-state controller in an incremental fashion that allows more fine-grained control of problem-solving, and it focuses computation where it is most likely to improve the value of the starting belief state. Testing on a wider range of examples using an improved implementation of the algorithm is planned to determine how far this approach may extend the range of POMDPs that can be solved by algorithms that use an exact piecewise linear and convex representation of the value function.

# 5   Conclusion

We have presented two related algorithms – an improved policy iteration algorithm and a new heuristic search algorithm – that solve infinite-horizon POMDPs by searching in a policy space of finite-state controllers.

Representation of a policy as a finite-state controller has a number of advantages. An optimal policy for a POMDP is sometimes equivalent to a finite-state controller, and when it is not, it can be approximated arbitrarily closely by a finite-state controller. Evaluation of a finite-state controller is straightforward and its value function is piecewise linear and convex. A finite-state controller can also be easier to understand than a policy represented (either explicitly or implicitly) as a mapping from regions of belief space to actions, and it can be executed without maintaining a belief state at run-time.

The bottleneck of both value iteration and policy iteration for POMDPs is the dynamic-programming update; Littman et al. (1995) prove that its worst-case complexity is exponential in the number of actions, observations, and vectors in the current value function. Policy iteration is faster than value iteration because it requires fewer iterations of the dynamic-programming update to converge to $\epsilon$-optimality. Because heuristic search can improve a finite-state controller without performing the dynamic-programming update, there is some reason to believe it may outperform policy iteration. Because the finite-state controller it finds optimizes the value of a starting belief state, and not the value of every possible belief state, it is usually smaller than the controller found by dynamic programming. Heuristic search also focuses computational effort on regions of the search space, or belief space, where improvement of the finite-state controller is most likely.

The heuristic search algorithm described here combines two areas of research on POMDPs that have developed indepedently. On the one hand, it draws from work on exact algorithms for POMDPs that use dynamic programming and a piecewise linear and convex representation of the value function (*e.g.*, Smallwood & Sondik 1973; Sondik 1978; Cassandra *et al.* 1994; Cassandra *et al.* 1997). On the other, it draws from work on approximation algorithms for POMDPs that perform forward search from a starting belief state, including work on computing bounds for the fringe nodes of a search tree (*e.g.*, Satia & Lave 1973; Larsen 1989; Washington 1996, 1997; Hauskrecht 1997). In the past, heuristic search has been used to find a solution that takes the form of a tree that grows as the depth of the search increases. The contribution of this paper is to show that heuristic search can find a compact finite-state controller (containing cycles) that describes infinite-horizon behavior.

## Acknowledgments.

Thanks to Shlomo Zilberstein and an anonymous reviewer for helpful comments and to Tony Cassandra and Milos Hauskrecht for making available their test problems. Support for this work was provided in part by the National Science Foundation under grants IRI-9624992, IRI-9634938 and INT-9612092.

## References

Brafman, R.I. (1997). A heuristic variable grid solution method for POMDPs. In *Proceedings of the Fifteenth National Conference on Artificial Intelligence*, 727–733. AAAI Press/The MIT Press.

Cassandra, A; Kaelbling, L.P.; and Littman, M.L. 1994. Acting Optimally in Partially Observable Stochastic Domains. In *Proceedings of the Twelfth National Conference on Artificial Intelligence*, 1023–1028. AAAI Press/The MIT Press.

Cassandra, A.; Littman, M.L.; and Zhang, N.L. 1997. Incremental pruning: A simple, fast, exact algorithm for partially observable Markov decision processes. In *Proceedings of the Thirteenth Annual Conference on Uncertainty in Artificial Intelligence*, 54-61. Morgan Kaufmann Publishers.

Chakrabarti, P.P; Ghose, S.; Acharya, A.; and de-Sarkar, S.C. 1990. Heuristic search in restricted memory. *Artificial Intelligence* 41:197–221.

Hansen, E.A. 1998a. An improved policy iteration algorithm for partially observable MDPs. In *Advances in Neural Information Processing Systems 10*. In press.

Hansen, E.A. 1998b. *Finite-Memory Control of Partially Observable Systems*. Ph.D. Diss., Department of Computer Science, University of Massachusetts at Amherst.

Hauskrecht, M. 1997. Incremental methods for computing bounds in partially observable Markov decision processes. In *Proceedings of the Fifteenth National Conference on Artificial Intelligence*, 734–739. AAAI Press/The MIT Press.

Kaelbling, L.P.; Littman, M.L.; and Cassandra, A.R. 1996. Planning and acting in partially observable stochastic domains. Computer Science Technical Report CS-96-08, Brown University.

Larsen, J.B. 1989. *A Decision Tree Approach to Maintaining a Deteriorating Physical System*. PhD thesis, University of Texas at Austin.

Littman, M.L.; Cassandra, A.R.; and Kaebling, L.P. 1995. Efficient dynamic-programming updates in partially observable Markov decision processes. Computer Science Technical Report CS-95-19, Brown University.

Satia, J.K. and Lave, R.E. 1973. Markovian Decision Processes with Probabilistic Observation of States. *Management Science* 20(1):1–13.

Smallwood, R.D. and Sondik, E.J. 1973. The optimal control of partially observable Markov processes over a finite horizon. *Operations Research* 21:1071-1088.

Sondik, E.J. 1978. The optimal control of partially observable Markov processes over the infinite horizon: Discounted costs. *Operations Research* 26:282-304.

Washington, R. 1996. Incremental Markov-model planning. In *Proceedings of TAI-96, Eighth IEEE International Conference on Tools with Artificial Intelligence*, 41–47.

Washington, R. 1997. BI-POMDP: Bounded, incremental partially-observable Markov-model planning. In *Proceedings of the Fourth European Conference on Planning*.